\documentclass{article}
\usepackage{collas2025_conference,times}

\usepackage{amsmath,amsfonts,bm}

\def\eqref#1{equation~\ref{#1}}

\def\1{\bm{1}}

\DeclareMathAlphabet{\mathsfit}{\encodingdefault}{\sfdefault}{m}{sl}
\SetMathAlphabet{\mathsfit}{bold}{\encodingdefault}{\sfdefault}{bx}{n}

\preprintcopy

\usepackage[T1]{fontenc}
\usepackage[utf8]{inputenc}
\usepackage{graphicx}
\graphicspath{{./images/}{./Graphs/}}
\usepackage{amsmath,amssymb,amsthm}
\usepackage{algorithm}
\usepackage{algpseudocode}
\usepackage{tikz}
\usetikzlibrary{arrows.meta,positioning,shapes,calc,fit}
\usepackage{float}
\usepackage{placeins}
\usepackage{wrapfig}
\usepackage{comment}
\usepackage{xcolor}
\usepackage{rotating}
\usepackage{csquotes}
\usepackage{appendix}
\usepackage{booktabs}
\usepackage{url}

\usepackage[hidelinks]{hyperref}
\hypersetup{
    pdftitle={Adjustment Speed as a Safety Constraint for Nonstationary Reinforcement Learning},
    pdfauthor={Timothy Tomashevskiy},
    pdfsubject={Reinforcement learning under nonstationarity and adaptive safety constraints},
    pdfkeywords={safe reinforcement learning, nonstationary reinforcement learning, adjustment speed, adaptation feasibility}
}

\newcommand{\parencite}[1]{\citep{#1}}
\newcommand{\textcite}[1]{\citet{#1}}

\newtheorem{lemma}{Lemma}
\newtheorem{proposition}{Proposition}

\title{Adjustment Speed as a Safety Constraint for Nonstationary Reinforcement Learning}
\author{ Timothy Tomashevskiy \\ Department of Computing and Software\\ McMaster University\\ Hamilton, Ontario, Canada\\ \texttt{ttm207@gmail.com} }

\begin{document}
\maketitle

\begin{abstract}
Ensuring safety in reinforcement learning under nonstationarity requires determining whether the learning system can adapt within the required recovery horizon under forecasted changes in the environment. Here, adjustment speed refers to the learning system's ability to safely adapt to forecasted environmental change. Existing approaches to safe reinforcement learning typically assume stationary environments and therefore do not explicitly consider adaptation speed under nonstationarity as a safety concern. However, when environments evolve over time, the rate at which the agent adapts to changing dynamics becomes critical for maintaining safety.

In this work, we propose adjustment speed as a safety constraint for nonstationary reinforcement learning. The key idea is to define safety in terms of adaptation feasibility: future states or regions may become unsafe not because they are inherently dangerous, but because the adaptation required to remain safe lies outside the learning system's calibrated recovery envelope. We formalize this relationship using representation learning and context forecasts to estimate environmental change and compare it with the learning system's achievable adaptation capacity.

The comparison between required adaptation and achievable adaptation is used to proactively regulate behavior when adaptation is predicted to become infeasible. Operationally, the framework tightens the admissible action set and activates an action-level shield when predicted adaptation demand exceeds the learning system's achievable adaptation capacity.

Empirical evaluation in a nonstationary driving environment shows that the proposed approach primarily reduces violations in short-horizon windows aligned to context changes. The ablations further show that shielding is more conservative for peak- and tail-risk suppression, while optimization-level adjustment provides additional reductions in short-horizon switch-conditioned violations.

These results support the view that safety under nonstationarity depends on adaptation feasibility: when forecasted environmental change occurs faster than the learning system can safely adapt, proactive safety intervention becomes necessary.

\end{abstract}
\section{Introduction}

Reinforcement learning (RL) has achieved strong empirical success in sequential decision-making problems such as robotics, autonomous driving, and adaptive control \parencite{mnih2015dqn,lillicrap2016ddpg}. However, many real-world deployment settings are inherently nonstationary: traffic patterns evolve, interacting agents change their behavior, sensor conditions drift, and task dynamics may vary over time. Such distribution shifts violate the stationarity assumptions under which most RL methods are developed and may substantially degrade both performance and safety.

This challenge is particularly acute in safety-critical settings. A policy that is safe under previously observed conditions may become unsafe after an environmental shift, even before any significant degradation in long-run performance is observed.
In practice, failures often arise not because the agent can never adapt, but because it cannot recover safe behavior within the required horizon. 
Immediately after a change in environment dynamics, the system may evolve before the policy has adapted to the new dynamics. 
During this transient phase, the agent continues to execute actions that may be suboptimal or unsafe under the new conditions. 
Adaptation after a distribution shift involves recovery of safe behavior, and delayed safety recovery can lead to bursts of unsafe behavior.

Most existing approaches to safe reinforcement learning do not explicitly address this issue. Classical methods based on constrained Markov decision processes, fixed safety specifications, or reward-penalty formulations are typically designed for stationary or slowly varying settings \parencite{altman1999constrained,achiam2017constrained,tessler2019reward}. At the same time, work on nonstationary, continual, and context-aware reinforcement learning has largely focused on adaptation and performance recovery rather than on maintaining safety throughout the transition itself \parencite{hallak2015contextual,finn2017model,kirkpatrick2017ewc}. Consequently, an important question remains insufficiently addressed: \emph{when is safe adaptation itself feasible under nonstationarity?}

In this paper, we argue that safety under nonstationarity should be understood through the relationship between the adaptation required by forecasted environmental change and the adaptation achievable by the learning system. The issue is not the physical speed of the agent, nor merely whether the policy can eventually adapt, but whether it can recover safe behavior within the required horizon as future context shifts unfold. When forecasted environmental change occurs faster than the learning system can safely adapt, transient unsafe behavior may become unavoidable unless additional protective mechanisms are activated.

This observation motivates the central concept of this work: \emph{adjustment speed as a safety constraint}. The goal is not to maximize adaptation speed itself, but to use adaptation feasibility as a criterion for future state admissibility. Predicted shifts in latent context induce a required adaptation demand; if that demand exceeds the learning system's achievable adaptation capacity, future states or regions affected by the shift are treated as unsafe for ordinary adaptation and require stronger safety intervention.

\begin{figure}[!t]
\centering
\vspace{-0.5em}
\begin{tikzpicture}[scale=0.78]
    \draw[->, thick] (0,0) -- (6.8,0) node[right] {Time};
    \draw[->, thick] (0,0) -- (0,4.5) node[above] {Change magnitude};
    \draw[blue, thick, dashed]
        plot[smooth] coordinates {(0,0.4) (1,0.9) (2,1.5) (3,2.0) (4,2.5) (5,3.0)};
    \draw[red, thick]
        plot[smooth] coordinates {(0,0.5) (1,1.1) (2,1.7) (3,2.6) (4,3.2) (5,3.9)};
    \fill[red!15]
        (2.8,2.0) -- (3.0,2.6) -- (4.0,3.2) -- (5.0,3.9)
        -- (5.0,3.0) -- (4.0,2.5) -- (3.0,2.0) -- cycle;
    \node[red] at (5.55,3.85) {$\Delta_E$};
    \node[blue] at (5.55,3.05) {$\Delta_A$};
    \node at (1.8,1.15) {\scriptsize Safe adaptation};
    \node at (4.15,3.15) {\scriptsize Unsafe region};
\end{tikzpicture}
\vspace{-0.5em}

\caption{Conceptual illustration of adaptation feasibility. The red curve denotes the required adaptation rate induced by forecasted environmental change, while the blue dashed curve denotes the learning system's achievable safe adaptation rate. The shaded region marks adaptation infeasibility, where environmental change occurs faster than the learning system can safely adapt. In this regime, ordinary learning may be too slow to maintain safety, motivating proactive intervention through constraint tightening and shielding.}

\label{fig:adaptation_feasibility}
\vspace{-1.2 em}
\end{figure}
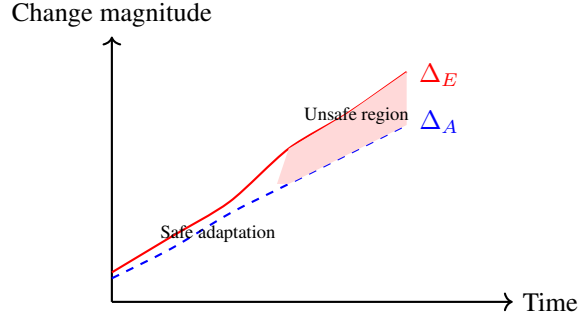

To implement this idea, we introduce \emph{Adjustment Speed as a Safety Constraint in Nonstationary Reinforcement Learning} (ASASC-NS). 
The framework learns a latent context representation of the current environment, predicts context evolution over a short adaptation horizon, estimates the adaptation demand induced by the predicted shift, and compares this quantity with the agent's empirically calibrated recovery capacity. When the required adaptation is predicted to exceed achievable safe adaptation, the framework proactively intervenes by tightening safety constraints and activating an action-level shielding mechanism.

We provide a conditional theoretical characterization of the relationship between predicted context shift, calibrated recovery capacity, and safety intervention. Empirically, the resulting mechanism primarily improves safety in short-horizon windows aligned to context changes, while the ablation clarifies that shielding is more conservative for peak- and tail-risk suppression. More broadly, this work advances safe continual reinforcement learning by shifting the focus from static safety specifications to \emph{feasible safe adaptation}.
\vspace{-0.5em}
\paragraph{Contributions.}
This paper makes four contributions. 
First, it introduces \emph{adjustment speed} as a safety constraint for reinforcement learning under nonstationarity and formalizes safety in terms of \emph{adaptation feasibility}, defined by whether the adaptation required by forecasted environmental change remains within the calibrated safe-recovery envelope for the required horizon. 
Second, it proposes a proactive framework (ASASC-NS) that combines context representation, context forecasting, adaptation-demand estimation, and action-level shielding to anticipate regimes in which safe adaptation may become infeasible. 
Third, it provides a conditional theoretical characterization of how environmental change and calibrated recovery capacity affect violation risk under distribution shift. 
Fourth, it empirically shows reduced safety violations in short-horizon windows aligned to context changes and analyzes how optimization-level adjustment and action-level shielding affect switch-conditioned, peak, and tail-risk metrics.

\section{Related Work}

Safe reinforcement learning is commonly studied in constrained Markov decision processes (CMDPs), where safety is enforced through explicit cost or risk constraints \parencite{altman1999constrained}. Representative approaches include constrained policy optimization \parencite{achiam2017constrained}, Lagrangian formulations \parencite{tessler2019reward}, and Lyapunov-based methods \parencite{chow2018lyapunov,berkenkamp2017safe}. These methods provide principled safety mechanisms, but typically assume stationary or slowly varying environments.

A second line of work enforces safety through shielding or safety filtering, where unsafe actions are modified before execution \parencite{alshiekh2018safe,dalal2018safe}. Hybrid approaches combine learned policies with control-theoretic safety filters \parencite{wabersich2021}. While effective, these mechanisms are largely reactive: they act after unsafe behavior is proposed rather than anticipating when changing conditions make safe adaptation difficult.

Research on nonstationary and continual reinforcement learning addresses changing environments through contextual MDPs \parencite{hallak2015contextual}, meta-learning \parencite{finn2017model}, continual learning and regularization against forgetting \parencite{kirkpatrick2017ewc}, and explicit modeling of context dynamics \parencite{xie2020lilac,nagabandi2019learning}. Recent work on plasticity loss in deep continual learning and continual deep RL shows that adaptation capacity itself can degrade or vary over time \parencite{dohare2024loss,abbas2023loss,juliani2024study}. Recent work has also begun to study safety under nonstationarity through context- or uncertainty-aware constraints \parencite{chen2021context,ding2023provably,wen2022improved}. However, existing methods generally do not provide an explicit operational criterion for determining when environmental change outpaces the agent's ability to adapt safely. We therefore treat $C_{\mathrm{adapt}}$ as a calibrated recovery threshold rather than a complete neural-plasticity model.

In contrast, we model safety under nonstationarity through \emph{adaptation feasibility}. The proposed framework predicts when adaptation becomes infeasible, identifies unsafe state regions that should be avoided, and uses this signal to guide policy adaptation and tighten action-level shielding. This shifts the focus from reactive correction to proactive avoidance of unsafe state regions.

\section{Preliminaries}

We consider reinforcement learning in nonstationary environments modeled as a time-dependent Markov decision process $\mathcal{M}_t=(\mathcal{S},\mathcal{A},P_t,R_t)$, where the transition kernel and reward function may vary over time. To capture this variability, we assume the environment is governed by a latent context variable $z_t\in\mathcal{Z}$ such that transitions depend on $(s_t,a_t,z_t)$.

Safety is defined through an unsafe set $\mathcal{U}\subset\mathcal{S}$, with violations indicated by
\[
v_t=\mathbf{1}\{s_t\in\mathcal{U}\}.
\]
The objective is to minimize violations while maintaining task performance.

When the context changes, a policy learned under previous dynamics may no longer remain safe or effective. If the rate of environmental change exceeds the agent's ability to adapt, transient safety violations may occur. This motivates mechanisms that explicitly account for the relationship between environmental change and adaptation capacity.

\section{Method}
\label{sec:method}

This section introduces \emph{Adjustment Speed as a Safety Constraint in Nonstationary Reinforcement Learning} (ASASC-NS). The method monitors predicted environmental change and asks whether the adaptation required to remain safe lies within the calibrated recovery envelope for the required horizon. When the predicted adaptation demand exceeds the learning system's achievable adaptation capacity, ASASC-NS uses the feasibility signal for optimization-level adjustment and for tightening the admissible action set through shielding.

\subsection{Context-Dependent Nonstationary MDP}
\label{subsec:context_mdp}

We model nonstationarity through a context-dependent MDP
\begin{equation}
M_{\kappa}=(\mathcal S,\mathcal A,P_{\kappa},r_{\kappa},c_{\kappa},\gamma),
\label{eq:context_mdp}
\end{equation}
where the latent context $\kappa$ indexes the transition dynamics, reward, and safety cost. The state and action spaces are shared across contexts, but the same state-action pair may induce a different next-state distribution when the context changes. In the driving domain, changes in traffic density, surrounding-vehicle speed, aggressiveness, and observation noise therefore affect $P_{\kappa}(s'\mid s,a)$ and $c_{\kappa}(s,a)$ rather than requiring a new state space. Safety is represented by a cost or risk signal $c_{\kappa}(s,a)$, and an action is admissible when its predicted risk is below the current threshold.

\subsection{Context Extraction and Prediction}
\label{subsec:context_extraction_prediction}

The context encoder is not a full model-based RL planner. It is a representation module used to estimate the current operating regime from recent transition evidence. Let
\begin{equation}
h_t=\{(s_i,a_i,s_{i+1})\}_{i=t-m}^{t}
\label{eq:transition_window}
\end{equation}
be a sliding window of recent transitions. The encoder maps this window to a latent context estimate
\begin{equation}
\hat{\kappa}_t=f_{\theta}(h_t).
\label{eq:context_encoder_window}
\end{equation}
In implementation, $f_{\theta}$ can be a recurrent encoder or Transformer over transition tuples. It is trained with an auxiliary context-conditioned next-state prediction loss,
\begin{equation}
\mathcal L_{\mathrm{ctx}}(\theta,\psi)=
\sum_{(s_i,a_i,s_{i+1})\in h_t}
\ell\!\left(s_{i+1},\hat p_{\psi}(s_i,a_i,\hat{\kappa}_t)\right)
+\lambda_{\mathrm{cons}}\mathcal L_{\mathrm{cons}},
\label{eq:ctx_loss}
\end{equation}
where $\mathcal L_{\mathrm{cons}}$ encourages windows from the same regime to produce similar context embeddings. This objective makes $\hat{\kappa}_t$ predictive of regime-level dynamics relevant to safety, without using the model for full model-based control.

A predictor forecasts the near-future context over a fixed short horizon $\Delta$:
\begin{equation}
\hat{\kappa}_{t+\Delta}=g_{\phi}(\hat{\kappa}_{t-L+1:t},\Delta).
\label{eq:context_predictor}
\end{equation}
In implementation, $g_\phi$ is a recurrent forecaster trained on observed context sequences using prediction error between the forecasted embedding $\hat\kappa_{t+\Delta}$ and the embedding later extracted from the realized transition window. The phrase ``near future'' therefore denotes this fixed prediction horizon. In the reported implementation we use $\Delta=10$ context-update steps, chosen as a fixed short horizon for forecasting near-term context evolution used by the adaptation-feasibility monitor.

\subsection{Adaptation Demand, Capacity, and Feasibility}
\label{subsec:adaptation_feasibility}

The predicted adaptation demand is the magnitude of the forecasted context displacement:
\begin{equation}
A_t=\left\|\hat{\kappa}_{t+\Delta}-\hat{\kappa}_t\right\|_2.
\label{eq:adaptation_demand_operational}
\end{equation}
Thus, $A_t$ measures predicted regime shift in the learned context space, not raw observation change. Since $\Delta$ is fixed, using $A_t$ is equivalent to using $A_t/\Delta$ for ranking the adaptation speed required over that horizon.

The adaptation capacity $C_{\mathrm{adapt}}$ is an empirical recovery-capacity threshold expressed in the same latent-distance units as $A_t$. It is calibrated from training rollouts as a conservative quantile of context displacements after which the agent recovers below a target violation level within a recovery horizon $H_{\mathrm{rec}}$:
\begin{equation}
C_{\mathrm{adapt}}=
\mathrm{Quantile}_{q}\!\left(
\{A_t:\bar v_{t:t+H_{\mathrm{rec}}}\le \eta\}
\right),
\label{eq:capacity_calibration}
\end{equation}
where $\bar v_{t:t+H_{\mathrm{rec}}}$ is the average violation indicator after the shift and $\eta$ is a tolerated recovery-level violation rate. The quantile $q$ is selected on calibration rollouts: smaller values make the trigger more conservative, while larger values allow larger shifts before intervention. This quantity should be interpreted as an operational safe-recovery threshold, not as a complete measure of neural plasticity.

\paragraph{Relation to plasticity.}
We do not claim that $C_{\mathrm{adapt}}$ is a complete measure of neural plasticity or parameter-level adaptability. Instead, it is an operational recovery-capacity threshold calibrated from observed post-shift learning behavior. This distinction separates our safety trigger from broader questions of plasticity loss in deep continual learning: ASASC-NS uses recovery evidence to decide when safety constraints should become more conservative, rather than attempting to characterize which network weights remain plastic.
The feasibility ratio is then
\begin{equation}
\rho_t=\frac{A_t}{C_{\mathrm{adapt}}+\epsilon},
\label{eq:feasibility_ratio_operational}
\end{equation}
where $\epsilon>0$ avoids division by zero. Values $\rho_t\le 1$ indicate shifts within the calibrated recovery envelope, while $\rho_t>1$ indicates predicted adaptation demand exceeding observed safe recovery capacity.

\subsection{Optimization-Level Adjustment}
\label{subsec:optimization_adjustment}

The feasibility signal can be used before action shielding as an optimization-level safety regulator. In the DQN instantiation, the reward used in the temporal-difference target is adjusted by a context-dependent safety penalty,
\begin{equation}
r_t^{\mathrm{AS}}=r_t-\beta_{\mathrm{AS}}\max(0,\rho_t-1)\,\hat c(s_t,a_t;\hat\kappa_t),
\label{eq:as_adjusted_reward}
\end{equation}
where $\beta_{\mathrm{AS}}\ge 0$ controls the strength of the adjustment. The resulting DQN loss is
\begin{equation}
\mathcal L_{\mathrm{DQN}}^{\mathrm{AS}}=
\left(Q_\omega(s_t,a_t)-\left[r_t^{\mathrm{AS}}+\gamma\max_{a'}Q_{\bar\omega}(s_{t+1},a')\right]\right)^2 .
\label{eq:as_dqn_loss}
\end{equation}
The \emph{Adj-only} ablation uses this optimization-level penalty without action filtering. The \emph{Shield-only} ablation sets $\beta_{\mathrm{AS}}=0$ and uses the same feasibility signal only to construct the admissible action set. The \emph{Full} method uses both mechanisms. This separation makes clear which part of the method affects learning updates and which part affects runtime action selection.

\subsection{Constraint Tightening and Action Shielding}
\label{subsec:tightening_shielding}

ASASC-NS converts adaptation infeasibility into a state-action safety intervention. The safety threshold is tightened when $\rho_t>1$:
\begin{equation}
\tau_t^{\mathrm{AS}}=\tau_0-\lambda \max(0,\rho_t-1),
\label{eq:tightened_threshold}
\end{equation}
where $\tau_0$ is the nominal threshold and $\lambda$ controls tightening strength. The admissible action set is
\begin{equation}
\mathcal A_{\mathrm{safe}}(s_t)=
\{a\in\mathcal A:\hat c(s_t,a;\hat\kappa_t)\le \tau_t^{\mathrm{AS}}\}.
\label{eq:admissible_action_set}
\end{equation}
In the driving experiments, $\hat c$ is computed from interpretable safety diagnostics such as collision or overlap risk, minimum gap, and time-to-collision. An action is unsafe if the predicted safety cost violates the tightened threshold. If the DQN action is inadmissible, the shield replaces it with the lowest-risk admissible action; if no admissible action is available, a conservative fallback action is executed.

\subsection{Algorithm Overview}
\label{subsec:algorithm_overview}

At each decision step, ASASC-NS updates the transition window, estimates the current context, predicts the future context, computes $A_t$ and $\rho_t$, and then selects an action subject to the tightened admissible set. The mechanism is proactive because intervention is triggered by predicted adaptation infeasibility rather than by an already observed safety violation.

\begin{algorithm}[t]
\caption{ASASC-NS decision step}
\label{alg:asasc_ns}
\begin{algorithmic}[1]
\State Update transition window $h_t=\{(s_i,a_i,s_{i+1})\}_{i=t-m}^{t}$
\State Estimate context $\hat\kappa_t=f_\theta(h_t)$
\State Forecast context $\hat\kappa_{t+\Delta}=g_\phi(\hat\kappa_{t-L+1:t},\Delta)$
\State Compute $A_t=\|\hat\kappa_{t+\Delta}-\hat\kappa_t\|_2$
\State Compute $\rho_t=A_t/(C_{\mathrm{adapt}}+\epsilon)$
\State Set $\tau_t^{\mathrm{AS}}=\tau_0-\lambda\max(0,\rho_t-1)$
\State Obtain policy action $a_t^{\mathrm{RL}}\sim \pi(\cdot\mid s_t)$
\If{$\hat c(s_t,a_t^{\mathrm{RL}};\hat\kappa_t)\le \tau_t^{\mathrm{AS}}$}
    \State Execute $a_t=a_t^{\mathrm{RL}}$
\Else
    \State Execute $a_t=\arg\min_{a\in\mathcal A_{\mathrm{safe}}(s_t)}\hat c(s_t,a;\hat\kappa_t)$, or fallback if $\mathcal A_{\mathrm{safe}}(s_t)=\emptyset$
\EndIf
\State Observe $s_{t+1}$ and, for adjustment variants, update DQN with $r_t^{\mathrm{AS}}$ from Eq.~\ref{eq:as_adjusted_reward}
\end{algorithmic}
\end{algorithm}
\section{Experiments}
\label{sec:experiments}

\begingroup
\setlength{\textfloatsep}{6pt plus 1pt minus 2pt}
\setlength{\floatsep}{6pt plus 1pt minus 2pt}
\setlength{\intextsep}{6pt plus 1pt minus 2pt}
\setlength{\abovecaptionskip}{3pt}
\setlength{\belowcaptionskip}{0pt}

We evaluate whether the proposed method reduces transient safety violations under nonstationarity and analyze how its optimization-level and shielding-level instantiations affect switch-aligned safety metrics.

\subsection{Experimental Setup}
\label{subsec:exp_setup}

\paragraph{Experimental instantiation.}
In \texttt{merge-v0}, nonstationarity is implemented as switching among traffic regimes that differ in traffic density, surrounding-vehicle speed, interaction aggressiveness, and observation noise. These regime changes modify the context-dependent transition distribution $P_{\kappa}(s'\mid s,a)$: for the same ego state and action, the next state may differ because surrounding vehicles move and interact differently across regimes. Thus, the experiments evaluate dynamics-level nonstationarity rather than merely introducing new states.

Safety is evaluated using collision and proximity-based diagnostics. A state-action pair is treated as unsafe when it is predicted to produce collision or overlap, insufficient front/rear gap, or time-to-collision below a safety threshold. The shield evaluates the discrete candidate actions available to the DQN policy using the predicted safety cost $\hat c(s_t,a;\hat\kappa_t)$. If the policy-selected action violates the current threshold, the shield replaces it with the lowest-risk admissible action; if no admissible action exists, a conservative fallback action is executed.

The ablations compare two operational uses of the same adaptation-feasibility signal. The adjustment-only variant uses this signal through the reward adjustment in Eq.~\ref{eq:as_adjusted_reward}, while the shield-only variant uses it to update a context-dependent reach-avoid safety specification and filter unsafe actions. The full method combines both uses. We report early-violation, peak-risk, tail-violation, and intervention-related diagnostics to distinguish transient-safety improvement from excessive conservatism.

\paragraph{Environment and nonstationarity.}
We evaluate on \texttt{merge-v0} from \texttt{highway-env}, augmented with a Markov context-switching mechanism to induce controlled nonstationarity. At each timestep, the environment remains in the current context with probability $p_{\text{stay}}$ and switches otherwise. Unless otherwise stated, we use a strong nonstationarity setting with $p_{\text{stay}}=0.5$, which induces frequent regime changes.

\paragraph{Agent and training.}
All methods use the same DQN architecture, optimizer, replay settings, and training horizon of 20k timesteps, differing only in the safety mechanism. We report results across $N=10$ random seeds. All runs use identical environment settings and the same context-switching process for a given seed.

\paragraph{Methods and ablations.}
We compare the following variants:
\begin{itemize}
\item \textbf{Baseline}: DQN without the adaptation-feasibility signal, without adjustment-speed-based optimization, and without action filtering.
\item \textbf{Adj-only}: DQN using the adaptation-feasibility signal to modify the optimization criterion, but without action filtering.
\item \textbf{Shield-only}: DQN using the same adaptation-feasibility signal to update a context-dependent reach-avoid safety specification and filter unsafe actions, but without the optimization-level adjustment.
\item \textbf{Full}: DQN combining optimization-level adjustment with context-dependent reach-avoid shielding.
\end{itemize}

\paragraph{Adjustment-speed activation and unsafe actions.}
The adaptation-feasibility signal is activated when the feasibility ratio satisfies $\rho_t > 1$. Unsafe actions are actions whose predicted safety cost exceeds the current context-dependent threshold. In the driving environment, this cost is computed from collision or overlap indicators and proximity diagnostics such as minimum gap and time-to-collision. When the policy proposes an unsafe action, the shield selects the lowest-risk admissible action or a conservative fallback if no admissible action is available.

\paragraph{Logging for switch-aligned analysis.}
To isolate transient failures caused by distribution shifts, we log per-step:
(i) the current context id $c_t$,
(ii) a switch flag $\mathbb{1}\{c_t\neq c_{t-1}\}$,
(iii) the violation indicator $v_t\in\{0,1\}$,
and optionally
(iv) a shield intervention flag.
This enables time-since-switch hazard curves and switch-aligned metrics.

\subsection{Safety Metrics}
\label{subsec:metrics}

Our goal is to reduce safety failures in short-horizon windows aligned to regime changes, not merely aggregate violation rates. We therefore report three complementary metrics computed from $v_t\in\{0,1\}$.

\paragraph{PeakRisk.}
We define the maximum rolling violation rate with window $W$:
\begin{equation}
\textsc{PeakRisk}=\max_t \frac{1}{W}\sum_{k=t-W+1}^{t} v_k.
\label{Peak_Risk_Rolling}
\end{equation}
We use $W=1000$ unless otherwise stated.

\paragraph{Switch-aligned EarlyViol and TailViol.}
Let $\{\tau_k\}_{k=1}^{K}$ be the timesteps at which a context switch occurs. For each switch time $\tau_k$, define an early window of length $H_{\text{early}}$ and a tail window $[H_{\text{tail-start}},H_{\text{tail-end}})$ measured after the switch. We compute:
\begin{equation}
\textsc{EarlyViol}=\frac{1}{K}\sum_{k=1}^{K}\frac{1}{H_{\text{early}}}
\sum_{t=\tau_k}^{\tau_k+H_{\text{early}}-1} v_t,
\label{Early+Violation}
\end{equation}
\begin{equation}
\textsc{TailViol}=\frac{1}{K}\sum_{k=1}^{K}\frac{1}{H_{\text{tail}}}
\sum_{t=\tau_k+H_{\text{tail-start}}}^{\tau_k+H_{\text{tail-end}}-1} v_t,
\quad H_{\text{tail}}=H_{\text{tail-end}}-H_{\text{tail-start}}.
\label{TailViolation}
\end{equation}
In our experiments we set $H_{\text{early}}=1000$, $H_{\text{tail-start}}=3000$, and $H_{\text{tail-end}}=5000$. These window sizes define short- and longer-horizon switch-conditioned summaries under the chosen switching frequency. Because additional switches may occur within either window, the metrics characterize behavior conditioned on a switch event rather than recovery from an isolated switch.

\paragraph{Reporting.}
For all scalar metrics we report mean $\pm$ 95\% confidence intervals across seeds.

\subsection{Main Results Under Strong Nonstationarity}
\label{subsec:main_results}

Table~\ref{tab:main_ablation} reports safety metrics under frequent context switching with $p_{\text{stay}}=0.5$. The full method achieves the lowest \textsc{EarlyViol}, indicating improved safety during the immediate adaptation phase, while the shield-only instantiation achieves the lowest \textsc{PeakRisk} and \textsc{TailViol} in this implementation.

\begin{table}[!htbp]
\centering
\caption{Safety metrics under frequent context switching ($p_{\text{stay}}=0.5$). We report mean $\pm$ 95\% CI across $N=10$ seeds. Lower is better for all safety metrics.}
\label{tab:main_ablation}
\begin{tabular}{lccc}
\hline
Method & EarlyViol $\downarrow$ & PeakRisk $\downarrow$ & TailViol $\downarrow$ \\
\hline
Baseline & 0.0690 $\pm$ 0.0069 & 0.0840 $\pm$ 0.0039 & 0.0698 $\pm$ 0.0054 \\
Adj-only & 0.0560 $\pm$ 0.0218 & 0.0415 $\pm$ 0.0267 & 0.0598 $\pm$ 0.0054 \\
Shield-only & 0.0510 $\pm$ 0.0118 & 0.0335 $\pm$ 0.0167 & 0.0308 $\pm$ 0.0064 \\
Full & 0.0440 $\pm$ 0.0103 & 0.0535 $\pm$ 0.0098 & 0.0402 $\pm$ 0.0031 \\
\hline
\end{tabular}
\end{table}

\paragraph{Interpreting adjustment, shielding, and their combination.}
Both the adjustment and shielding variants use the same adaptation-feasibility signal derived from context-forecasted nonstationarity and calibrated recovery capacity. The adjustment-only variant uses this signal in the adjusted TD target from Eq.~\ref{eq:as_dqn_loss}, whereas the shield-only variant uses it to construct an updatable context-dependent reach-avoid safety specification. The full method combines both interventions. Because the optimization-level adjustment and action-level shield influence the same policy behavior, their effects need not be additive; implementation-level interactions can make the combined method perform better on short-horizon switch-conditioned violations while the shield-only variant remains stronger on some peak-risk metrics. Thus, the ablation should be interpreted as comparing two operational uses of adaptation feasibility, not as showing that the shield is independent of the proposed mechanism.

\paragraph{Transient safety improvements.}
The largest gains appear in \textsc{EarlyViol}, which is consistent with the purpose of the adjustment-speed mechanism: to tighten safety specifically when predicted environmental change exceeds adaptation feasibility.

\paragraph{Worst-case burst behavior.}
\textsc{PeakRisk} captures rare but high-risk violation bursts. In this implementation, the shield-only variant attains the lowest peak value, suggesting that the reach-avoid shielding instantiation is especially effective at suppressing burst risk. The full method remains lower than the baseline, but does not dominate shield-only on \textsc{PeakRisk}. This is consistent with interaction effects between optimization-level adjustment and action-level shielding, rather than with a failure of adaptation-feasibility information itself.

\paragraph{Steady-state behavior.}
\textsc{TailViol} reflects violation rates in a longer-horizon window following each recorded switch. The lower tail violation rates of shield-only and full suggest that adaptation-feasibility-guided safety interventions remain useful beyond the short-horizon switch-conditioned window. However, the main advantage of the full method is concentrated in early adaptation rather than in uniformly dominating every instantiation on all metrics.

\subsection{Ablation Summary}
\label{subsec:ablation}

The ablation results suggest that the two uses of the adaptation-feasibility signal play different roles. Optimization-level adjustment contributes most directly to improved early adaptation safety, as reflected in the reduction of \textsc{EarlyViol}, while reach-avoid shielding is particularly effective at suppressing high-risk bursts and lowering later violation levels. Their combination yields the strongest early-phase safety performance, but the shield-only instantiation remains stronger on \textsc{PeakRisk} and \textsc{TailViol} in this implementation.

\paragraph{Scope of the empirical claim.}
The results should be interpreted as evidence for targeted transient-safety improvement rather than uniform dominance of the combined method over each component. Shield-only uses the same adaptation-feasibility mechanism to update the reach-avoid safety specification and may achieve lower peak or tail violations when shielding dominates the safety behavior. The full method combines shielding and optimization-level adjustment; interactions between them may improve early safety without improving every peak- or tail-risk metric.

\subsection{Nonstationarity Sweep}
\label{subsec:pstay_sweep}

We sweep switching severity by varying $p_{\text{stay}}\in\{0.50,0.70,0.85,0.95,0.99\}$ and report \textsc{EarlyViol} in Figure~\ref{fig:pstay_sweep}. Lower $p_{\text{stay}}$ corresponds to more frequent regime switches. The purpose of this sweep is not to claim monotonic behavior in every finite-sample setting, but to test whether the feasibility gate remains useful across switching frequencies. Across the evaluated settings, the proposed method maintains lower short-horizon switch-conditioned violation rates than the unconstrained baseline, although the size of the gap varies with switching frequency and confidence intervals.

\subsection{Time-Since-Switch Hazard Analysis}
\label{subsec:hazard}

To analyze switch-conditioned safety bursts, we align trajectories by the most recent context switch. Let $\Delta_t=t-\max\{\tau_k:\tau_k\le t\}$ denote time since the last switch.

Violations occur as sparse, burst-like events rather than as a continuous signal, which results in spiky hazard curves even after aggregation. Accordingly, the analysis focuses on relative differences in violation frequency and intensity across methods, particularly at small values of time since the most recent switch.

Figure~\ref{fig:switch_hazard} plots the empirical violation probability as a function of $\Delta_t$. The baseline exhibits elevated violation probability at several values of time since the most recent switch. The full method shows lower violation levels over portions of the short-horizon switch-conditioned interval, consistent with proactively tightening constraints when predicted change exceeds adaptation feasibility.

\vspace{-0.5em}
\begin{figure}[!htbp]
\centering
\begin{minipage}[t]{0.48\linewidth}
    \centering
    \includegraphics[width=\linewidth]{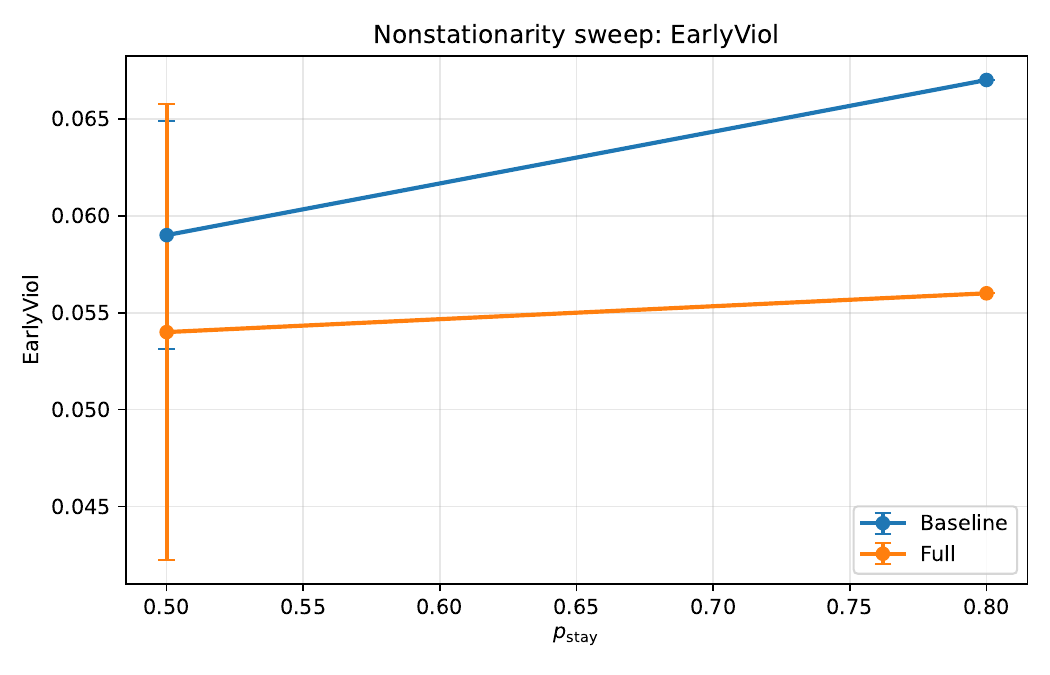}
    \caption{\textbf{Effect of nonstationarity on transient safety.}
    Early-phase violation rate (\textsc{EarlyViol}) as a function of the context-persistence probability $p_{\text{stay}}$. Lower values of $p_{\text{stay}}$ correspond to more frequent context switches and stronger nonstationarity. The proposed method maintains lower violation rates across the displayed switching regimes, although the improvement is not monotonic and the gap varies across settings.}
    \label{fig:pstay_sweep}
\end{minipage}
\hfill
\begin{minipage}[t]{0.48\linewidth}
    \centering
    \includegraphics[width=\linewidth]{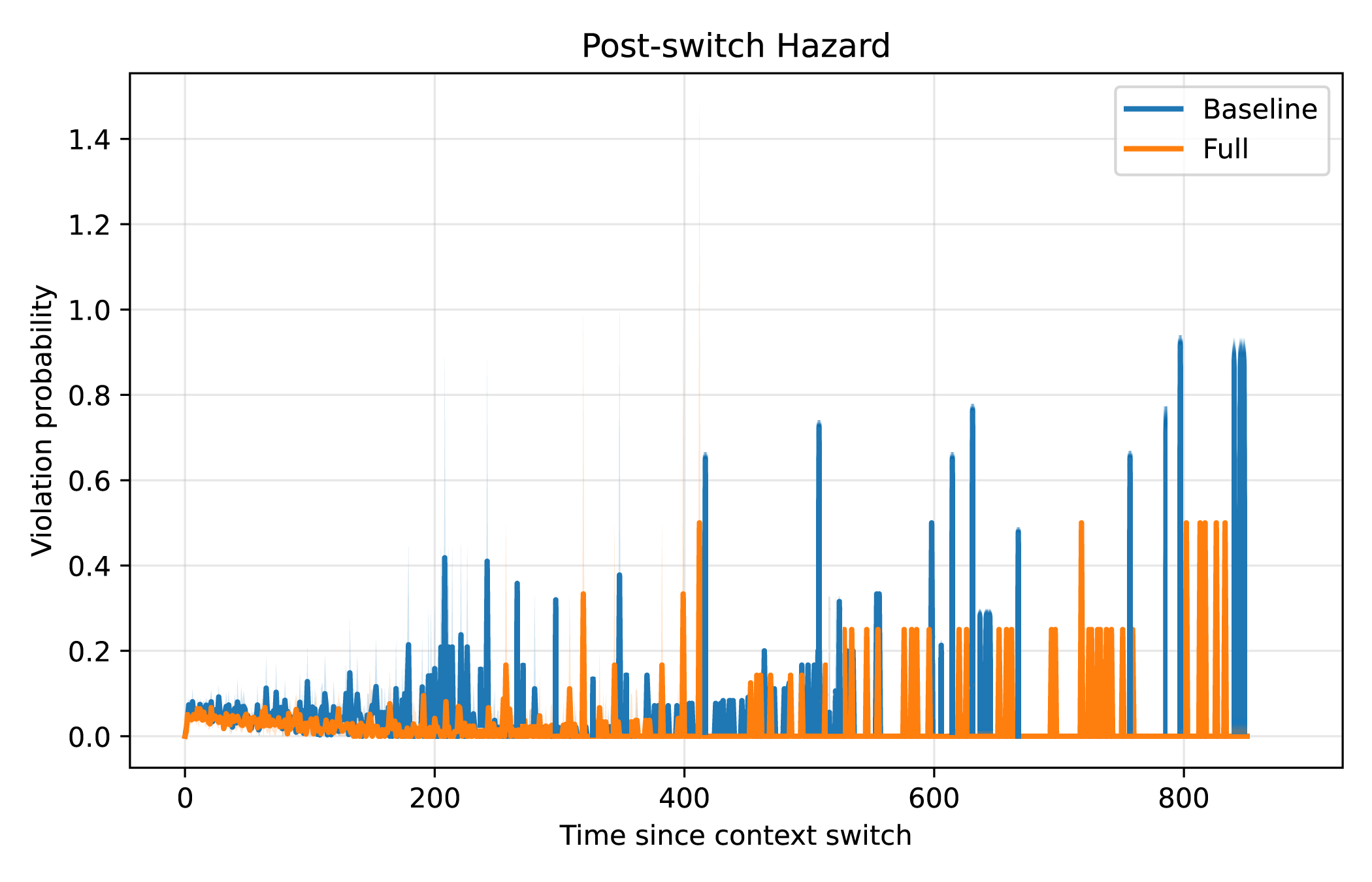}
    \caption{\textbf{Transient safety after distribution shift.}
    Empirical violation probability $\Pr(v_t=1 \mid \Delta_t)$ as a function of time since the most recent context switch. Despite visible variability due to the bursty nature of violations, the full method shows reduced violation levels over portions of the short-horizon switch-conditioned interval.}
    \label{fig:switch_hazard}
\end{minipage}
\vspace{-0.5em}
\end{figure}
\vspace{-0.75em}

\subsection{Additional Visualizations}
\label{subsec:viz}

To complement the aggregate safety metrics, we visualize the temporal evolution of violations during training.
Figure~\ref{fig:rolling_viols} shows the rolling violation rate, highlighting transient bursts that occur when the agent encounters unfamiliar regimes. Consistent with the main results, the full method shows lower rolling violation levels during early adaptation.
Figure~\ref{fig:cumulative_viols} reports the cumulative mean violation rate over the training horizon. The separation between curves during earlier training is consistent with lower observed violation rates for the full method, while the later reduction in separation indicates smaller differences during later training.

\vspace{-0.5em}
\begin{figure}[!htbp]
\centering
\begin{minipage}[t]{0.48\linewidth}
    \centering
    \includegraphics[width=\linewidth]{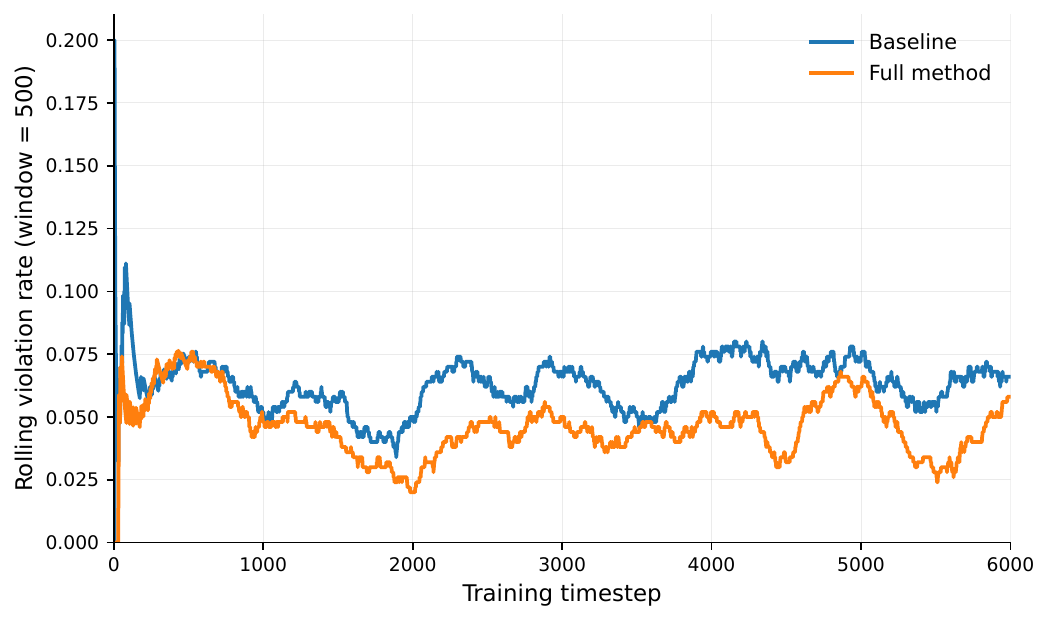}
    \caption{Rolling violation rate during early training under $p_{\text{stay}}=0.5$. The full method shows lower rolling violation levels during adaptation to unfamiliar regimes.}
    \label{fig:rolling_viols}
\end{minipage}
\hfill
\begin{minipage}[t]{0.48\linewidth}
    \centering
    \includegraphics[width=\linewidth]{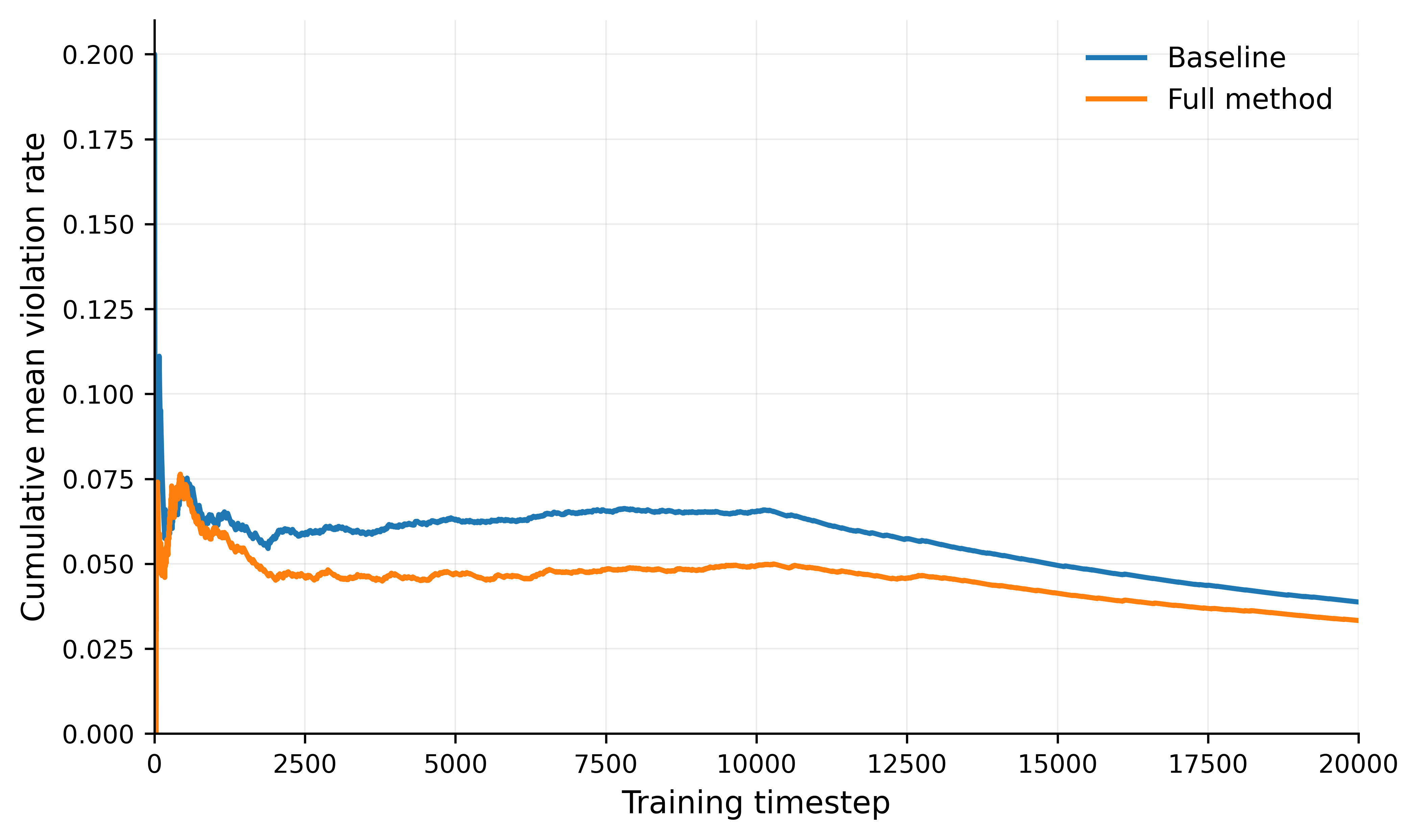}
    \caption{Cumulative mean violation rate during training. Early-phase safety improvements are visible, while the curves become closer later in training.}
    \label{fig:cumulative_viols}
\end{minipage}
\vspace{-0.5em}
\end{figure}
\vspace{-0.75em}

\FloatBarrier

\paragraph{Scope and limitations.}
This work does not aim to characterize catastrophic forgetting or neural plasticity at the parameter level. Instead, we study adaptation feasibility as a safety condition: given a predicted context shift, the question is whether the required adaptation remains within the calibrated recovery envelope for the specified horizon. If the predicted adaptation demand exceeds the calibrated recovery capacity, then ordinary adaptation is treated as unsafe and stronger intervention is required, such as tighter constraints, shielding, or context-dependent policy adjustment. Thus, $C_{\mathrm{adapt}}$ is not a general plasticity measure; it is an operational safety threshold used to decide when predicted nonstationarity makes otherwise reachable regions unsafe under the agent's current adaptation capability. The present evaluation focuses on the safety diagnostics directly targeted by the proposed adaptation-feasibility mechanism and does not provide a complete task-performance analysis. Broader task-return reporting and additional environments remain important directions for follow-up evaluation.

\endgroup
\section{Conclusion}

This paper introduced adaptation feasibility as a safety principle for reinforcement learning under nonstationarity. The objective is not to maximize adaptation speed itself, but to determine whether the adaptation required by forecasted environmental change remains within the calibrated recovery envelope for the required horizon. Future states or regions may become unsafe when ordinary adaptation cannot keep pace with the required adaptation demand.

The proposed framework operationalizes adaptation feasibility. Nonstationarity is represented through a context-dependent MDP, current context is estimated from transition windows, future context is predicted over a short horizon, and adaptation demand is measured as the predicted displacement in latent context space. Adaptation capacity is treated as an empirical recovery threshold, not as a complete model of neural plasticity. When the feasibility ratio exceeds one, the method tightens the safety threshold and restricts the admissible action set through shielding.

The experiments show that this mechanism primarily improves safety in short-horizon windows aligned to context changes, which is the intended target of the adjustment-speed constraint. The ablation also shows that the shielding-only instantiation can be more conservative for peak and tail risk. Thus, the empirical contribution should be interpreted as evidence that adaptation-feasibility monitoring helps identify high-risk transition periods, while optimization-level adjustment and reach-avoid shielding provide complementary uses of the same safety signal.

Overall, this work positions \emph{adaptation feasibility} as a useful principle for safe reinforcement learning under nonstationarity. Safety depends not only on whether a constraint exists, but also on whether the required adaptation remains within the calibrated recovery envelope for the required horizon under forecasted environmental change. Future work will extend this formulation to richer forms of nonstationarity, stronger capacity estimators, and broader evaluations of the trade-off between intervention frequency, conservatism, and safety.

{\small
\bibliographystyle{collas2025_conference}
\bibliography{bib}
}

\appendix
\section{Conditional Theoretical Characterization}
\label{app:theory}

This section provides a conditional theoretical characterization of adaptation feasibility under nonstationary environments. 
The analysis formalizes the intuition that predicted context shifts can increase violation risk when they exceed the agent's calibrated recovery capacity. 
The proposed adjustment-speed mechanism mitigates this effect locally by tightening the admissible action set when adaptation is predicted to become unsafe.

\subsection{Adaptation Demand and Capacity}
\label{app:adaptation_demand_capacity}

Let $\hat{\kappa}_t$ denote the latent context estimated from the recent transition window $h_t$, and let $\hat{\kappa}_{t+\Delta}$ denote the predicted future context over horizon $\Delta$. 
Following Section~\ref{subsec:context_extraction_prediction}, the context estimate is obtained as
\begin{equation}
\hat{\kappa}_t = f_{\theta}(h_t),
\qquad
h_t=\{(s_i,a_i,s_{i+1})\}_{i=t-m}^{t},
\end{equation}
and the future context is predicted as
\begin{equation}
\hat{\kappa}_{t+\Delta}
=
g_{\phi}(\hat{\kappa}_{t-L+1:t},\Delta).
\end{equation}
The predicted adaptation demand is defined as
\begin{equation}
A_t =
\left\|
\hat{\kappa}_{t+\Delta}-\hat{\kappa}_t
\right\|_2 .
\label{eq:appendix_adaptation_demand}
\end{equation}
Thus, $A_t$ measures predicted displacement in the learned context space rather than raw observation change.

Let $C_{\mathrm{adapt}}$ denote the calibrated recovery-capacity threshold introduced in Section~\ref{subsec:adaptation_feasibility}. 
We treat this quantity as an operational threshold estimated from observed post-shift safety recovery, not as a complete measure of neural plasticity. 
The adaptation-feasibility ratio is
\begin{equation}
\rho_t =
\frac{A_t}{C_{\mathrm{adapt}}+\epsilon},
\label{eq:appendix_feasibility_ratio}
\end{equation}
where $\epsilon>0$ avoids division by zero. 
Values $\rho_t\le 1$ correspond to shifts within the calibrated recovery envelope, while $\rho_t>1$ indicates that predicted adaptation demand exceeds observed safe recovery capacity.

\paragraph{Why a quantile threshold?}
The quantile calibration is not intended to estimate a universal biological or neural-plasticity constant. It is a validation-set rule for converting observed post-shift recovery behavior into an operational safety threshold. Choosing a lower quantile produces earlier interventions and more conservative behavior; choosing a higher quantile reduces intervention frequency but accepts larger predicted shifts before shielding. This makes $C_{\mathrm{adapt}}$ reproducible from logged rollouts and keeps it directly comparable to $A_t$.

\subsection{Violation Probability Under Context Shift}
\label{app:violation_shift}

We now formalize the relationship between context shift and safety violations.

\begin{lemma}[Violation probability under context shift]
Assume that the policy $\pi_t$ was trained or adapted under context $\kappa_t$ and that the environment transitions toward context $\kappa_{t+\Delta}$. 
Let
\[
\delta_t =
\|\kappa_{t+\Delta}-\kappa_t\|_2
\]
denote the context-shift magnitude. 
If the transition kernel and safety cost are Lipschitz continuous with respect to context, then the probability of entering an unsafe state satisfies
\[
\mathbb{P}(v_t=1)
\le
\alpha + L\delta_t,
\]
where $\alpha$ is the baseline violation probability under the original context and $L$ captures the sensitivity of the safety event to context change.
\end{lemma}

\paragraph{Proof sketch.}
Under Lipschitz continuity, a bounded context shift induces a bounded change in the transition distribution and safety cost. 
Because the policy was optimized or adapted under the original context, the shifted transition kernel may increase the probability of reaching unsafe states. 
Bounding this change by the context-shift magnitude yields the stated inequality.

\subsection{Adaptation Feasibility}
\label{app:adaptation_feasibility}

The previous lemma shows that violation probability can increase with context shift. 
However, the learning system may reduce this risk if it can adapt safely before entering high-risk state regions.

\begin{lemma}[Adaptation feasibility]
Suppose that the calibrated recovery threshold $C_{\mathrm{adapt}}$ upper-bounds the context shifts after which the agent has historically recovered below the tolerated violation level within the recovery horizon. 
If
\[
A_t \le C_{\mathrm{adapt}},
\]
then the predicted context shift lies within the observed safe-recovery envelope. 
If
\[
A_t > C_{\mathrm{adapt}},
\]
then the shift is outside this envelope and should be treated as adaptation-infeasible for safety control.
\end{lemma}

\paragraph{Proof sketch.}
By construction, $C_{\mathrm{adapt}}$ is calibrated from rollouts in which the agent recovered below a target violation level after context shifts of comparable magnitude. 
When $A_t \le C_{\mathrm{adapt}}$, the predicted shift falls within this empirical recovery envelope. 
When $A_t>C_{\mathrm{adapt}}$, the method cannot rely on ordinary policy adaptation alone to maintain safety and therefore activates stricter safety control.

\subsection{Effect of Constraint Tightening}
\label{app:tightening_effect}

When $\rho_t>1$, ASASC-NS tightens the safety threshold and restricts the admissible action set:
\begin{equation}
\tau_t^{\mathrm{AS}}
=
\tau_0-\lambda\max(0,\rho_t-1),
\label{eq:appendix_threshold}
\end{equation}
\begin{equation}
\mathcal A_{\mathrm{safe}}(s_t)
=
\{a\in\mathcal A:
\hat c(s_t,a;\hat\kappa_t)
\le
\tau_t^{\mathrm{AS}}\}.
\label{eq:appendix_safe_set}
\end{equation}

\begin{proposition}[Risk reduction under tightened admissibility]
Assume that the safety-cost estimator $\hat c$ ranks candidate actions consistently with their realized violation risk up to bounded estimation error. 
When $\rho_t>1$, replacing an inadmissible action with the lowest-risk admissible action weakly reduces the predicted one-step violation risk relative to executing the inadmissible policy action.
\end{proposition}

\paragraph{Proof sketch.}
If the policy-selected action violates the tightened threshold, then by definition it lies outside $\mathcal A_{\mathrm{safe}}(s_t)$. 
The shield selects an admissible action with lower predicted safety cost or executes a conservative fallback. 
Under bounded estimation error and consistent risk ranking, this replacement weakly reduces predicted violation risk. 
The guarantee is conditional on the quality of the safety-cost estimate and should be interpreted as a local admissibility guarantee rather than an unconditional hard-safety guarantee.

\subsection{Interpretation}
\label{app:theory_interpretation}

The theoretical results clarify the role of adaptation feasibility. 
Predicted context shifts can increase safety risk, and shifts outside the calibrated recovery envelope indicate that ordinary policy adaptation may be insufficient. 
The adjustment-speed mechanism responds by tightening admissibility and restricting actions in regions where safe adaptation is predicted to be infeasible.

These guarantees are intentionally conditional. 
They do not claim that $C_{\mathrm{adapt}}$ is a full neural-plasticity measure or that the combined method must dominate every shielding instantiation on every metric. 
Instead, they support the central claim of the paper: context-forecasted adaptation demand can be used as an operational safety signal for identifying high-risk transition periods and proactively restricting unsafe actions.

\section{Implementation Details}
\label{app:implementation_details}

This appendix gives the implementation details that are most relevant for reproducibility. It mirrors the notation in Section~\ref{sec:method} and avoids treating the context module as a full model-based RL planner.

\paragraph{RL backbone.}
All reported discrete-control experiments use DQN with the same network, optimizer, replay buffer, and training horizon across methods. The safety mechanisms are applied after the DQN proposes an action and before the action is executed. The policy update itself uses the same replay data and optimizer settings for all variants, except that the adjustment-only and full variants include the adjustment-speed regulation term described below.

\paragraph{Optimization-level adjustment.}
For the Adj-only and Full variants, the DQN update uses the adjusted reward in Eq.~\ref{eq:as_adjusted_reward}. The penalty is zero when $\rho_t\le 1$ and increases with both predicted adaptation infeasibility and predicted safety cost. This gives a concrete implementation of ``adjustment-speed-based optimization'' without action filtering. The Shield-only variant disables this term by setting $\beta_{\mathrm{AS}}=0$.

\paragraph{Context extraction.}
The context encoder receives a sliding transition window
\[
h_t=\{(s_i,a_i,s_{i+1})\}_{i=t-m}^{t},
\]
rather than a single state. In our implementation, each transition tuple is vectorized and passed through a small recurrent encoder. The resulting embedding $\hat\kappa_t=f_\theta(h_t)$ summarizes regime-level properties of the local dynamics, such as surrounding-vehicle density, relative speed pattern, aggressiveness, and observation noise. The encoder is trained with the auxiliary loss
\[
\mathcal L_{\mathrm{ctx}}=\sum_{(s_i,a_i,s_{i+1})\in h_t}\ell\!\left(s_{i+1},\hat p_\psi(s_i,a_i,\hat\kappa_t)\right)+\lambda_{\mathrm{cons}}\mathcal L_{\mathrm{cons}},
\]
where the prediction term encourages $\hat\kappa_t$ to encode transition-relevant information and the consistency term encourages windows from the same regime to produce similar embeddings. The learned transition predictor is used only to shape the representation; it is not used as a full planning model.

\paragraph{Context prediction and horizon.}
A recurrent predictor forecasts context from recent embeddings,
\[
\hat\kappa_{t+\Delta}=g_\phi(\hat\kappa_{t-L+1:t},\Delta).
\]
The horizon $\Delta=10$ context-update steps defines the fixed short interval over which near-term context evolution and the associated adaptation demand are forecast. Thus, ``near future'' refers to this specified prediction horizon rather than to an unspecified task horizon.

\paragraph{Adaptation demand and calibrated capacity.}
The adaptation demand is
\[
A_t=\|\hat\kappa_{t+\Delta}-\hat\kappa_t\|_2 .
\]
The capacity term is calibrated in the same units as $A_t$. Specifically, $C_{\mathrm{adapt}}$ is estimated from training rollouts as a high quantile of context displacements after which the agent recovers below a target violation level within a recovery horizon:
\[
C_{\mathrm{adapt}}=\mathrm{Quantile}_{q}\bigl(\{A_t:\bar v_{t:t+H_{\mathrm{rec}}}\le \eta\}\bigr).
\]
This makes $\rho_t=A_t/(C_{\mathrm{adapt}}+\epsilon)$ a dimensionless feasibility ratio. The quantile $q$ controls the conservatism of the trigger and is selected on calibration rollouts together with $\eta$. $C_{\mathrm{adapt}}$ should be interpreted as an empirical safe-recovery threshold, not as a parameter-level model of plasticity.

\paragraph{Constraint tightening and shielding.}
When $\rho_t>1$, the threshold is tightened according to
\[
\tau_t^{\mathrm{AS}}=\tau_0-\lambda\max(0,\rho_t-1).
\]
For each discrete candidate action, the shield computes a safety-cost estimate $\hat c(s_t,a;\hat\kappa_t)$ from collision or overlap risk, gap margins, and time-to-collision. The admissible set is
\[
\mathcal A_{\mathrm{safe}}(s_t)=\{a:\hat c(s_t,a;\hat\kappa_t)\le \tau_t^{\mathrm{AS}}\}.
\]
If the DQN action is admissible, it is executed. Otherwise, the shield executes the lowest-risk admissible action. If no admissible action exists, a conservative fallback action is used.

\paragraph{Nonstationarity in \texttt{merge-v0}.}
The Markov context-switching process changes regime-level traffic parameters including density, surrounding-vehicle speed distribution, interaction aggressiveness, and observation noise. These changes affect the context-dependent transition kernel $P_\kappa(s'\mid s,a)$: the same ego state and action can lead to a different next-state distribution because surrounding vehicles behave differently under different contexts.

\paragraph{Logged quantities.}
For every run we log context id, switch indicators, violation indicators, predicted context displacement $A_t$, feasibility ratio $\rho_t$, tightened threshold $\tau_t^{\mathrm{AS}}$, and shield intervention indicators. These logs support the switch-aligned metrics, hazard curves, ablations, and intervention analysis.

\section{Additional Experimental and Calibration Details}
\label{app:additional_details}

This appendix provides supplementary implementation details. The main theoretical characterization is given in Appendix~\ref{app:theory}; we therefore avoid repeating a second, inconsistent proof or algorithm here.

\subsection{Calibration of the Feasibility Threshold}
\label{app:capacity_calibration_details}

The recovery-capacity threshold $C_{\mathrm{adapt}}$ is calibrated from training rollouts using the same latent-distance units as $A_t$. For each detected or induced context shift, we compute the predicted displacement $A_t$ and the post-shift average violation rate over a recovery horizon $H_{\mathrm{rec}}$. Shifts whose post-shift violation rate remains below the tolerated level $\eta$ are treated as empirically recoverable. We then set
\[
C_{\mathrm{adapt}}=\mathrm{Quantile}_{q}\bigl(\{A_t:\bar v_{t:t+H_{\mathrm{rec}}}\le \eta\}\bigr).
\]
This calibration is deliberately operational: it estimates the largest context displacement for which the current learning system has historically recovered safely. It does not attempt to identify which network weights remain plastic or to model catastrophic forgetting at the parameter level.

\subsection{Additional Algorithmic Detail}
\label{app:algorithmic_detail}

The deployed decision step is the same as Algorithm~\ref{alg:asasc_ns} in the main paper. The key implementation choices are:
\begin{itemize}
    \item context is inferred from transition windows $(s_t,a_t,s_{t+1})$, not from a single state;
    \item the context predictor uses $\Delta=10$ context-update steps as a fixed short forecasting horizon;
    \item $A_t$ and $C_{\mathrm{adapt}}$ are both measured in latent-context distance units;
    \item tightening means decreasing the admissible safety threshold $\tau_t^{\mathrm{AS}}$ as $\rho_t$ grows above one;
    \item shielding replaces inadmissible DQN actions with the lowest-risk admissible action or with a conservative fallback.
\end{itemize}

\subsection{Context-Prediction Diagnostic}
\label{app:context_prediction_diagnostic}

To address whether the latent predictor is meaningful, the implementation logs the realized forecast error
\[
e^{\mathrm{pred}}_t = \|\hat\kappa_{t+\Delta}-\hat\kappa^{\mathrm{obs}}_{t+\Delta}\|_2,
\]
where $\hat\kappa_{t+\Delta}$ is the predicted future context and $\hat\kappa^{\mathrm{obs}}_{t+\Delta}$ is the embedding extracted later from the realized transition window. This diagnostic is used to check that the adaptation-demand signal reflects actual context evolution. The main paper reports downstream safety metrics because the proposed constraint is evaluated by whether it reduces post-shift violations, not by context-prediction error alone.

\section{Hyperparameters}
\label{app:hyperparameters}

Table~\ref{tab:hyperparameters} summarizes the main hyperparameters used in the reported experiments. The safety-specific parameters are calibrated on training rollouts as described in Appendix~\ref{app:capacity_calibration_details}.

\begin{table}[H]
\centering
\caption{Key training and safety-monitoring hyperparameters.}
\label{tab:hyperparameters}
\begin{tabular}{lc}
\hline
Parameter & Value / selection rule \\
\hline
RL backbone & DQN \\
Learning rate & $10^{-4}$ \\
Discount factor & $0.99$ \\
Replay buffer size & $10^6$ \\
Batch size & $64$ \\
Training steps & $20{,}000$ \\
Strong-switching setting & $p_{\mathrm{stay}}=0.5$ \\
Context window length $m$ & validation-selected transition window \\
Context history length $L$ & validation-selected embedding history \\
Prediction horizon $\Delta$ & $10$ context-update steps \\
Recovery horizon $H_{\mathrm{rec}}$ & $H_{\text{early}}=1000$ steps \\
Capacity quantile $q$ & high quantile of recoverable shifts \\
Tolerance $\eta$ & validation-selected target violation level \\
Adjustment penalty $\beta_{\mathrm{AS}}$ & validation-selected; disabled for Shield-only \\
Tightening strength $\lambda$ & validation-selected safety-margin slope \\
Fallback action & conservative lowest-risk action \\
\hline
\end{tabular}
\end{table}

\end{document}